\def\paperID{18656}
\def\confName{CVPR}
\def\confYear{2025}

\def\paperTitle{ViQAgent: Zero-Shot Video Question Answering \\via Agent with Open-Vocabulary Grounding Validation}

\def\authorBlock{
    \textbf{Tony Montes \quad
    Fernando Lozano} \\ 
    The Department of Electrical and Electronics Engineering\\ Universidad de los Andes\\
    {\small \texttt{\{t.montes, flozano\}@uniandes.edu.co}}
}

\newif\ifreview \newcommand{\review}{\reviewtrue}
\newif\ifarxiv \newcommand{\arxiv}{\arxivtrue}
\newif\ifcamera 
\newif\ifrebuttal 

\arxiv 

\pdfoutput=1
\documentclass[10pt,twocolumn,letterpaper]{article}
\ifreview \usepackage[review]{cvpr} \fi
\ifarxiv \usepackage[pagenumbers]{cvpr} \fi
\ifrebuttal \usepackage[rebuttal]{cvpr} \fi
\ifcamera \usepackage{cvpr} \fi

\usepackage{graphicx}	
\usepackage{amsmath}	
\usepackage{amssymb}	
\usepackage{booktabs}
\usepackage{times}
\usepackage{microtype}
\usepackage{epsfig}
\usepackage[table,xcdraw,dvipsnames]{xcolor}
\usepackage{caption}
\usepackage{float}
\usepackage{placeins}
\usepackage{color, colortbl}
\usepackage{stfloats}
\usepackage{enumitem}
\usepackage{tabularx}
\usepackage{xstring}
\usepackage{multirow}
\usepackage{xspace}
\usepackage{url}
\usepackage{subcaption}
\usepackage{xcolor}
\usepackage[hang,flushmargin]{footmisc}

\usepackage{pifont}
\usepackage{algorithm}
\usepackage{algpseudocode}

\usepackage{listings}
\usepackage{tcolorbox}

\newtcolorbox[auto counter, number within=section]{mybox}[2][]{colback=blue!10!white,colframe=blue!80!black,
fonttitle=\bfseries, title=#2,#1}

\newtcolorbox[auto counter, number within=section]{promptbox}[2][]{colback=green!10!white,colframe=green!80!black,
fonttitle=\bfseries, title=Prompt: #2,#1}

\ifcamera \usepackage[accsupp]{axessibility} \fi

\newcommand{\cmark}{\ding{51}}%
\newcommand{\xmark}{\ding{55}}%

% ...

\ifarxiv  \fi

\newcommand{\R}[1]{{%
    \textbf{%
        \ifstrequal{#1}{1}{\textcolor{red}{R#1}}{%
        \ifstrequal{#1}{2}{\textcolor{blue}{R#1}}{%
        \ifstrequal{#1}{3}{\textcolor{magenta}{R#1}}{%
        \ifstrequal{#1}{4}{\textcolor{teal}{R#1}}{%
                           \textcolor{cyan}{R#1}%
        }}}}%
    }%
}}

\usepackage{xr-hyper}

\makeatletter
\newcommand*{\addFileDependency}[1]{
  \typeout{(#1)}
  \@addtofilelist{#1}
  \IfFileExists{#1}{}{\typeout{No file #1.}}
}

\makeatother
\newcommand*{\myexternaldocument}[1]{
    \externaldocument{#1}
    \addFileDependency{#1.tex}
    \addFileDependency{#1.aux}
}

\definecolor{cvprblue}{rgb}{0.21,0.49,0.74}
\usepackage[pagebackref,breaklinks,colorlinks,citecolor=cvprblue]{hyperref}
\usepackage[capitalize]{cleveref}
\crefname{section}{Sec.}{Secs.}
\crefname{table}{Table}{Tables}
\crefname{figure}{Fig.}{Figs.}

\ifarxiv \crefname{appendix}{App.}{Apps.}
\else \crefname{appendix}{Suppl.}{Suppls.} \fi

\unless\ifarxiv \myexternaldocument{_supplementary} \fi

\begin{document}

\title{\paperTitle}
\author{\authorBlock}
\maketitle

\begin{abstract} 

Recent advancements in Video Question Answering (VideoQA) have introduced LLM-based agents, modular frameworks, and procedural solutions, yielding promising results. These systems use dynamic agents and memory-based mechanisms to break down complex tasks and refine answers. However, significant improvements remain in tracking objects for grounding over time and decision-making based on reasoning to better align object references with language model outputs as newer models get better at both tasks. This work presents an LLM-brained agent for zero-shot Video Question Answering (VideoQA) that combines a Chain-of-Thought framework with grounding reasoning alongside YOLO-World to enhance object tracking and alignment. This approach establishes a new state-of-the-art in VideoQA and Video Understanding, showing enhanced performance on NExT-QA, iVQA, and ActivityNet-QA benchmarks. Our framework also enables cross-checking of grounding timeframes, improving accuracy and providing valuable support for verification and increased output reliability across multiple video domains. The code is available \href{https://github.com/t-montes/viqagent}{here}.

\end{abstract}

\begin{figure*}[t]
  \includegraphics[width=\linewidth]{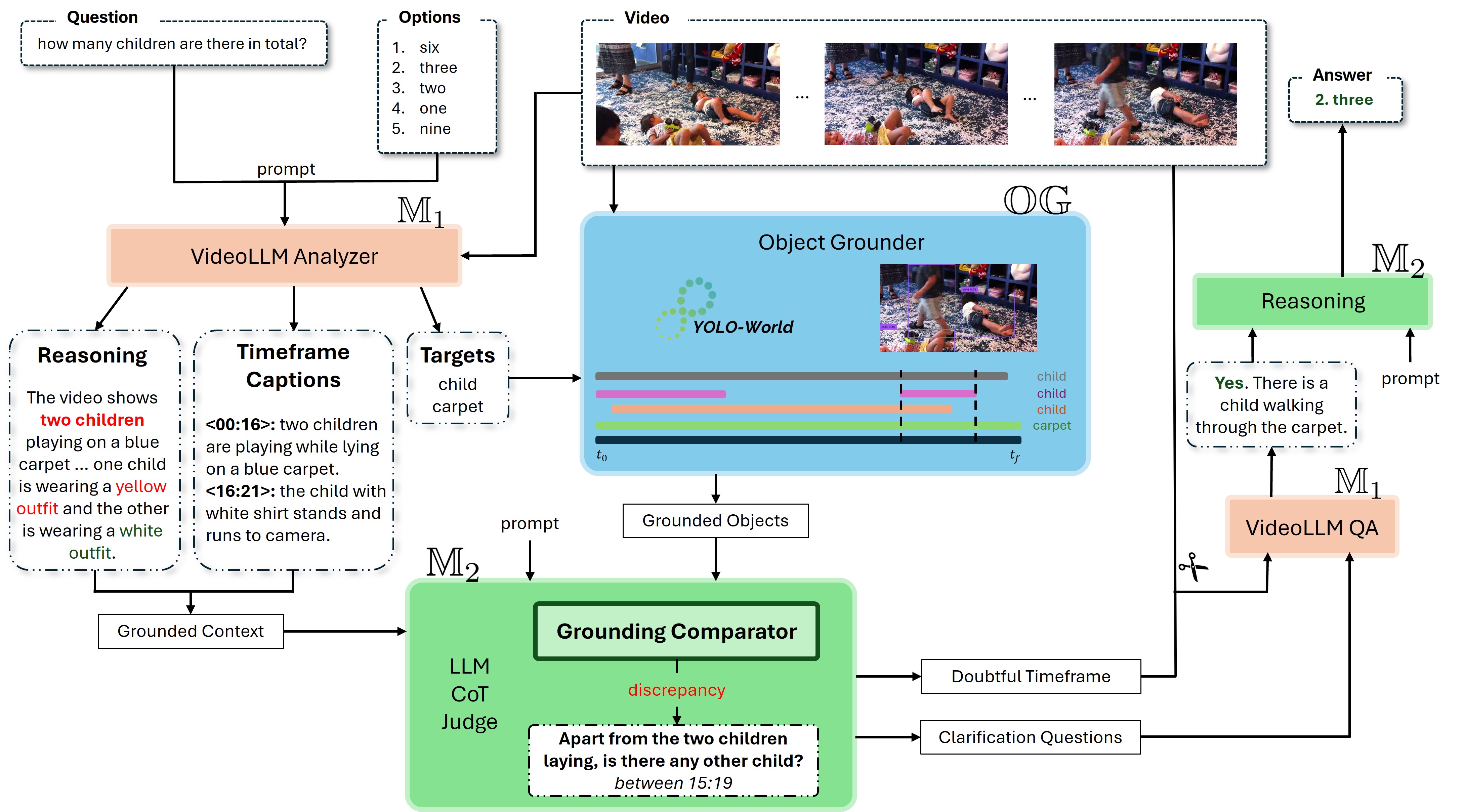} \hfill
  \caption {\textbf{An overview of our ViQAgent framework}. Through three main modules, we propose an agentic solution for the Video Question-Answering (VideoQA) task by taking advantage of most advanced VideoLLMs capabilities on first-sight zero-shot reasoning, timeframe captioning, and target identification {\color{BurntOrange} ($M_1$)}, and the open-vocabulary capabilities of YOLO-World to ground the given targets/objects in the video {\color{Cyan} ($OG$)} in specific parts of the video in between $t_0$ and $t_f$; to finally end with a Chain-of-Thoughts judgment and reasoning layer {\color{ForestGreen} $(M_2)$} that compares both the grounded context and grounded object detections to determine the confidence of the $M_1$ answer. In case of discrepancy, the CoT judge defines a set of clarification questions in specific timeframes that go through the VideoLLM again for specific short-ended question-answering. Finally, a reasoning layer takes these answers and the original question to produce a grounded and more accurate answer.}
  \label{fig:pic1}
\end{figure*}

\section{Introduction}
\label{sec:intro}

In the present year, the evolution of large language models (LLMs) \cite{gpt4, gemini15, mistral} and vision language models (VLMs) \cite{videollava, videollama, videochat, videochatgpt, blip2, pali, palix} has significantly advanced their ability in video understanding, particularly in the video question-answering (VideoQA) task, a significant challenge on computer vision, where the model is provided with a video and a related question \cite{vqallm, vqallm_2, vqa_grounding_long, vqa_long_movie} that they must answer as accurately as possible. These models are designed to analyze the visual and linguistic data of the video to generate answers based on semantics and dynamics. Despite recent progress, significant limitations still need to be addressed when addressing more complex videos, particularly those with dynamic context and extensive length. Current LLM-based solutions often struggle with reliably capturing content that is crucial for answering questions when scenes are complex or require a high level of contextual and sequential understanding. This highlights the need for more adaptable approaches that can respond to a broader range of video types and question complexities \cite{vqa_alignment, vqa_question_instructed, vqa_rag, vqa_reducing_hallucination}.

To address these challenges, agent-based strategies for VideoQA have recently emerged, employing modular reasoning blocks \cite{morevqa_modular_cot_reasoner, traveler_agent_cot_planner}, memory-based strategies \cite{videoagent_memory, morevqa_modular_cot_reasoner}. Procedural approaches \cite{zeroshot_procedural, traveler_agent_cot_planner}, as well as employing foundational vision-language models as tools for solving complex tasks and augmenting the context \cite{videoagent_long, zeroshot_procedural, videoofthought_cot_reasoner}. This enables models to track relevant content more effectively over time, which results especially useful for structuring tasks and reasoning through multiple steps, enabling more accurate question answering even when handling diverse and dynamic content. A key approach in these frameworks is video grounding \cite{vqa_grounding_long, cot_grounding_evaluator, morevqa_modular_cot_reasoner, videoofthought_cot_reasoner}, wherein specific video segments are identified as containing objects or events essential to answering the question. In this case, Grounding anchors target segments within the video, allowing the model to focus on relevant portions rather than attempting to analyze entire video sequences. These approaches are also increasingly incorporating Chain-of-Thought (CoT) reasoning. In this method, the model explicitly articulates step-by-step reasoning with purposes of evaluating the consistency \cite{cot_grounding_evaluator, cot_qa_evaluator, cot_vision_evaluator}, planning the steps to solve a task \cite{traveler_agent_cot_planner}, or just bringing a more accurate and argued response \cite{cot_reasoner, cot_vision_reasoner, videoofthought_cot_reasoner, morevqa_modular_cot_reasoner}. CoT reasoning can enhance certainty in the model's responses, improve interpretability by presenting intermediate reasoning that is understandable to humans, and validate model outputs against additional contextual information.

Inspired by these advances, we propose ViQAgent. This framework combines the capabilities of VideoLLMs and state-of-the-art vision-language foundation models to create a structured agent for VideoQA. ViQAgent employs a VideoLLM \cite{gemini15} to identify the key objects (\ie, targets) in the video that are relevant to the question, generating an initial understanding of the main elements and dynamics in a video and a preliminary answer based on observed video content. This initial answer (\ie, \textit{first-sight} response) includes a first attempt to ground it in time and provides a set of key timeframes with their corresponding captions in the video. In the next phase, ViQAgent utilizes YOLO-World \cite{yoloworld} to perform object detection based on this list of targets. Here, YOLO-World tracks occurrences of these predefined targets across video frames, returning an accurate timeline of their appearances, a task that sometimes VLMs tend to struggle with, when compared against each other \cite{gpt4v_yoloworld}. Unlike traditional object detectors with fixed categories, YOLO-World's tracking is customized by the VideoLLM’s selected open-vocabulary targets, allowing it to focus detection on the objects that are most pertinent to solving the question. The timeline created in this step adds a layer of precision, extending the initial detection with frame-by-frame data for each target object and the count of detected objects. In the final phase, the judgment and reasoning layer compares the initial response plus its grounded context, as well as the comprehensive object-tracking data, to determine whether the response is confident or not. This reasoning layer applies Chain-of-Thought (CoT) reasoning to combine both inputs, carefully validating the initial reaction against the grounded data from YOLO-World to produce an answer that is both accurate and substantiated by clear visual evidence. In case of an unconfident answer, a set of expressly framed questions are re-validated through the VideoLLM to provide a final answer then. The CoT approach in this output stage reinforces interpretability by tracing logical steps in reaching the answer and strengthens answer reliability by cross-checking the data from both sources.

Our approach provides multiple advantages as a zero-shot solution, requiring no specialized task-based fine-tuning and thus allowing easy adaptation to new scenarios and question types with just a dataset-specific subprompt. By combining the strengths of VideoLLMs for initial analysis and general video understanding and YOLO-World for open-vocabulary object detailed tracking, it benefits from the complementary capabilities of these models, handling a wide range of video-based questions with minimal configuration. Furthermore, the CoT mechanism contributes to a more nuanced and compelling cross-validation of the reasoning outputs. This grounding structure improves reliability and helps ensure that the content in the video firmly supports all answers. The open nature of ViQAgent’s vocabulary for target tracking - dynamically adapted by the VideoLLM layer based on the question - further enhances its flexibility and relevance for various VideoQA tasks.

Empirical results underscore the effectiveness of ViQAgent, as it consistently outperforms current zero-shot solutions across major benchmarks: NExT-QA \cite{nextqa} \cite{egoschema}, iVQA \cite{ivqa}, ActivityNet-QA \cite{activitynetqa}, and the Egoschema's open subset \cite{egoschema}, achieving up to a 4.4\% improvement in accuracy. This improvement sets a new state-of-the-art for VideoQA systems, especially in zero-shot scenarios where temporal and spatial reasoning capabilities are essential.

In summary, our contributions are as follows: \textbf{(1)} Implementation of ViQAgent, a zero-shot framework that integrates VideoLLMs and vision-language models for effective video grounding, object tracking, and question answering in VideoQA. \textbf{(2)} Enhanced interpretability through structured Chain-of-Thought reasoning and grounding outputs, providing interpretable insight into the intermediate steps that inform the final response of the model. \textbf{(3)} State-of-the-art zero-shot performance across NExT-QA, iVQA, and ActivityNet-QA benchmarks, highlighting ViQAgent's effectiveness in addressing complex temporal and spatial reasoning tasks with higher accuracy and adaptability.

\section{Related Work}
\label{sec:related}

\textbf{Video Question-Answering.} Video Question-Answering (VideoQA) has seen significant progress in recent years \cite{vqa_grounding_long, nllm_vqa_challenges, vqa_alignment, vqa_question_instructed, vqa_rag, vqa_reducing_hallucination, nllm_vqa_clip, nllm_grounding}, contributing advancements in both video understanding and natural language processing. Early VideoQA models focused on straightforward tasks, such as frame captioning \cite{nllm_vqa_clip} or simple event identification. Still, these approaches were often limited to fundamental interactions or small, static frameworks \cite{vqallm, vqallm_2}. Recent methods have introduced more sophisticated architectures that incorporate attention mechanisms \cite{vqa_alignment, vqa_question_instructed, vqa_rag, vqa_reducing_hallucination}, temporal modeling \cite{vqa_grounding_long, nllm_grounding}, and multi-modal transformers \cite{vqa_long_movie, videollama, blip2, videollava, videochat, videochatgpt} to handle complex questions over temporally extended video sequences. Benchmarks such as NExT-QA \cite{nextqa}, iVQA \cite{ivqa}, ActivityNet-QA \cite{activitynetqa}, and EgoSchema \cite{egoschema} have challenged models with complex queries that require contextual awareness, reasoning over sequential frames, and an understanding of nuanced interactions. Despite these advancements, current end-to-end approaches struggle with contextual continuity across frames and often require task-specific training to achieve high accuracy. Zero-shot solutions, while desirable, have had limited success in maintaining high generalizability and usually require fine-tuning to increase performance significantly. This highlights the need for more adaptable methods or vision-language tools to respond to diverse question types. ViQAgent seeks to address this gap by utilizing an agent-based approach that grounds key video elements in response to the question, adding a layer of interpretability that can better navigate complex VideoQA challenges in zero-shot scenarios.
\\
\textbf{LLM Modular Agents.} Modular agents built on large language models (LLMs) \cite{morevqa_modular_cot_reasoner, traveler_agent_cot_planner, videoagent_long, videoagent_memory, videoofthought_cot_reasoner, zeroshot_procedural} have gained traction as a solution for decomposing tasks into manageable components (\ie, modules), such as in MoReVQA \cite{morevqa_modular_cot_reasoner} and TraveLER \cite{traveler_agent_cot_planner}, enabling improved specialization and adaptability in complex problem-solving domains, including VideoQA. These agents use LLMs to orchestrate a series of modular reasoning blocks, each designed to address specific sub-tasks within a broader query. This strategy allows the model to segment complex questions, apply targeted reasoning processes, and sequentially consolidate insights into a final answer. Many modern modular agents also incorporate memory mechanisms, such as VideoAgent \cite{videoagent_memory}, TraveLER \cite{traveler_agent_cot_planner}, and MoReVQA \cite{morevqa_modular_cot_reasoner}. This approach enhances their ability to retain relevant contextual information over time. This is particularly useful in VideoQA, where questions often require remembering objects or events across multiple frames and reasoning about their changes over time. Furthermore, the integration of language-vision foundation models as tools within these modular agents has allowed for more accurate detection, such as in VideoAgent \cite{videoagent_long}, ProViQ \cite{zeroshot_procedural}, and MotionEpic \cite{videoofthought_cot_reasoner}. ViQAgent builds upon this approach by combining VideoLLMs \cite{gemini15} with YOLO-World \cite{yoloworld}, which adds targeted grounding capabilities to the agent's toolkit, specifically tracking object appearances and interactions as specified by the VideoLLM’s outputs. This allows for an additional layer of analysis in complex scenes, enabling the agent to consolidate multiple perspectives and enhance the accuracy and interpretability of its answers.
\\
\textbf{Chain-of-Thought Reasoning.} Chain-of-thought (CoT) reasoning has emerged as a promising approach in complex question-answering, allowing models to break down tasks into sequential, interpretable steps \cite{cot_dataset, cot_grounding_evaluator, cot_qa_evaluator, cot_reasoner, cot_vision_evaluator, cot_vision_reasoner, traveler_agent_cot_planner, videoofthought_cot_reasoner, morevqa_modular_cot_reasoner}. In VideoQA, where questions often require multistep reasoning to contextualize events, CoT enables models to articulate intermediate reasoning steps, improving the accuracy of the answer and enhancing the interpretability of the model’s outputs. By outputting a transparent sequence of reasoning steps, CoT makes it possible to validate each stage of the decision-making process \cite{cot_vision_reasoner, morevqa_modular_cot_reasoner}, adding a layer of reliability to model predictions \cite{cot_qa_evaluator, cot_grounding_evaluator, cot_vision_evaluator}, this is particularly observed on the VoT reasoning framework \cite{videoofthought_cot_reasoner}. Grounding, a related concept within CoT frameworks, involves identifying and tracking objects or actions relevant to a given question across frames \cite{nllm_grounding, vqa_grounding_long, cot_grounding_evaluator}. This process creates a temporal map of pertinent elements, providing a more structured basis for the CoT to build upon. ViQAgent leverages CoT by incorporating it within its final judgment and reasoning layer, using CoT steps to align the preliminary answer and grounded object tracking data from YOLO-World, ultimately refining the response based on these validated insights through several validation questions about them. This structured approach ensures that each reasoning step is clear and accessible, supporting the model’s interpretability and enhancing its overall robustness in VideoQA tasks.

\section{ViQAgent Framework}
\label{sec:method}

\begin{figure*}[t]
  \includegraphics[width=\linewidth]{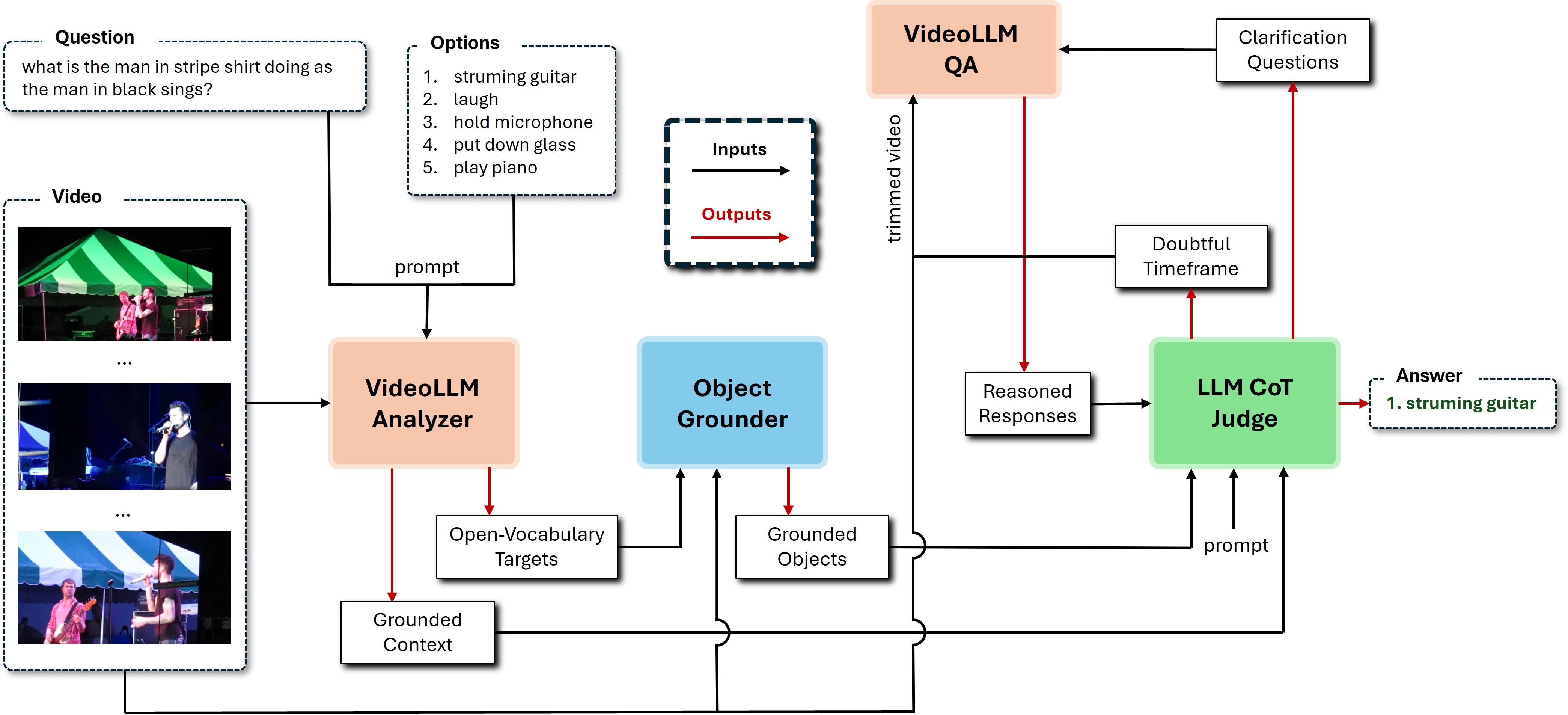} \hfill
  \caption {An outline of the black-boxed ViQAgent framework modules \textbf{inputs} and {\color{BrickRed} \textbf{outputs}}, and the intermediate representations, that allow to track and understand the final selected answer. The {\color{BurntOrange} ($M_1$)} \textbf{inputs} are the video and the question plus the answer options (namely \textit{prompt}). In contrast, the {\color{BrickRed} \textbf{outputs}} are the open-vocabulary targets, and the reasoning plus timeframe captions (namely \textit{Grounded Context}). The {\color{Cyan} ($OG$)} \textbf{inputs} are the targets and the video, and the {\color{BrickRed} \textbf{output}} is the object detection timeline (namely \textit{Grounded Objects}). Finally, the {\color{ForestGreen} $(M_2)$} first \textbf{receives} both ground responses and the prompt, then, if there seem to be inconsistencies, {\color{BrickRed} \textbf{returns}} a doubtful timeframe and a set of clarification questions to make to the VideoLLM from that specific timeframe. The answers are then \textbf{re-inputted} to {\color{BrickRed} \textbf{produce}} the final answer.}
  \label{fig:pic2}
\end{figure*}

In this section, we describe in detail the implementation details of each module of the ViQAgent framework (\cref{fig:pic2}), starting with a general overview and task definition (\cref{sec:method-overview}), module implementation details (\cref{sec:method-m1}, \cref{sec:method-og}, and \cref{sec:method-m2}), to finally end with an algorithm of the full implementation (\cref{sec:method-all}).

\subsection{Overview}
\label{sec:method-overview}

ViQAgent answers both open- and close-ended questions about video content using three interconnected modules. First, a VideoLLM-based module $M_1$ provides initial insights into the video's relevance to the question. Next, an open-vocabulary grounding module $OG$ detects and tracks detected targets across frames. Finally, a reasoning module $M_2$ validates, cross-checks, and refines answers using prior outputs, forming a cohesive chain-of-thought question-based reasoning process. ViQAgent thus identifies, tracks, and reasons over relevant video segments, generating accurate answers with minimal prior knowledge or domain-specific data.

In a VideoQA task, the model is given a question $Q$ and a video $V$, composed of $n$ frames $V = [v_1, ..., v_n]$, based on the video's frames-per-second ratio. For close-ended questions, the model also receives a set of answer options $A_{opt}$, ensuring that the output answer must be in that set $A \in A_{opt}$ \cite{morevqa_modular_cot_reasoner}. Only $Q$ and $V$ are provided for open-ended questions, though $A_{opt}$ is typically used to gauge answer similarity before the ground truth and the model's output. Our pipeline aims to solve the VideoQA task with an intermediate rationale for increasing interpretability. The task can be defined as follows, assuming a solution system $S$:
\begin{equation}
    S(V,Q+[A_{opt}]) \rightarrow A
\end{equation}

\subsection{VideoLLM Analyzer}
\label{sec:method-m1}

The analyzer module $M_1$, also referred to as the \textit{first-sight assessment}, employs a VideoLLM to provide a preliminary interpretation of the video for the question. Given the question $Q$, and $A_{opt}$, and the entire video $V$, this module generates three outputs:
\\
\textbf{1. Open-Vocabulary Targets:} The VideoLLM identifies a set of open-vocabulary targets $\{ T_1, T_2, ..., T_m \}$ essential to solving the question. These targets are derived from both the question context and the video content, forming an initial roadmap of entities or events likely to contribute to the answer.
\\
\textbf{2. Preliminary Answer and Reasoning:} Based on its analysis, the VideoLLM proposes an initial answer $A_1$ to the question, with a detailed rationale $R_1$ that articulates why this answer might be appropriate, and that will later aid in the validation process.
\\
\textbf{3. Scene-Segmented Timeframes:} The VideoLLM segments the video into discrete scenes based on changes it identifies as relevant to the question and generates a caption describing in detail what happens in that scene. These segmented timeframes serve as markers for different key moments in the video, where one significant scene ends and another begins, establishing a structured temporal foundation for subsequent grounding and reasoning steps. These segments can be modeled as a set of pairs of timeframe-caption $\{ (t_i,t_f): c \}$.

\begin{figure}[t]
  \includegraphics[width=\linewidth]{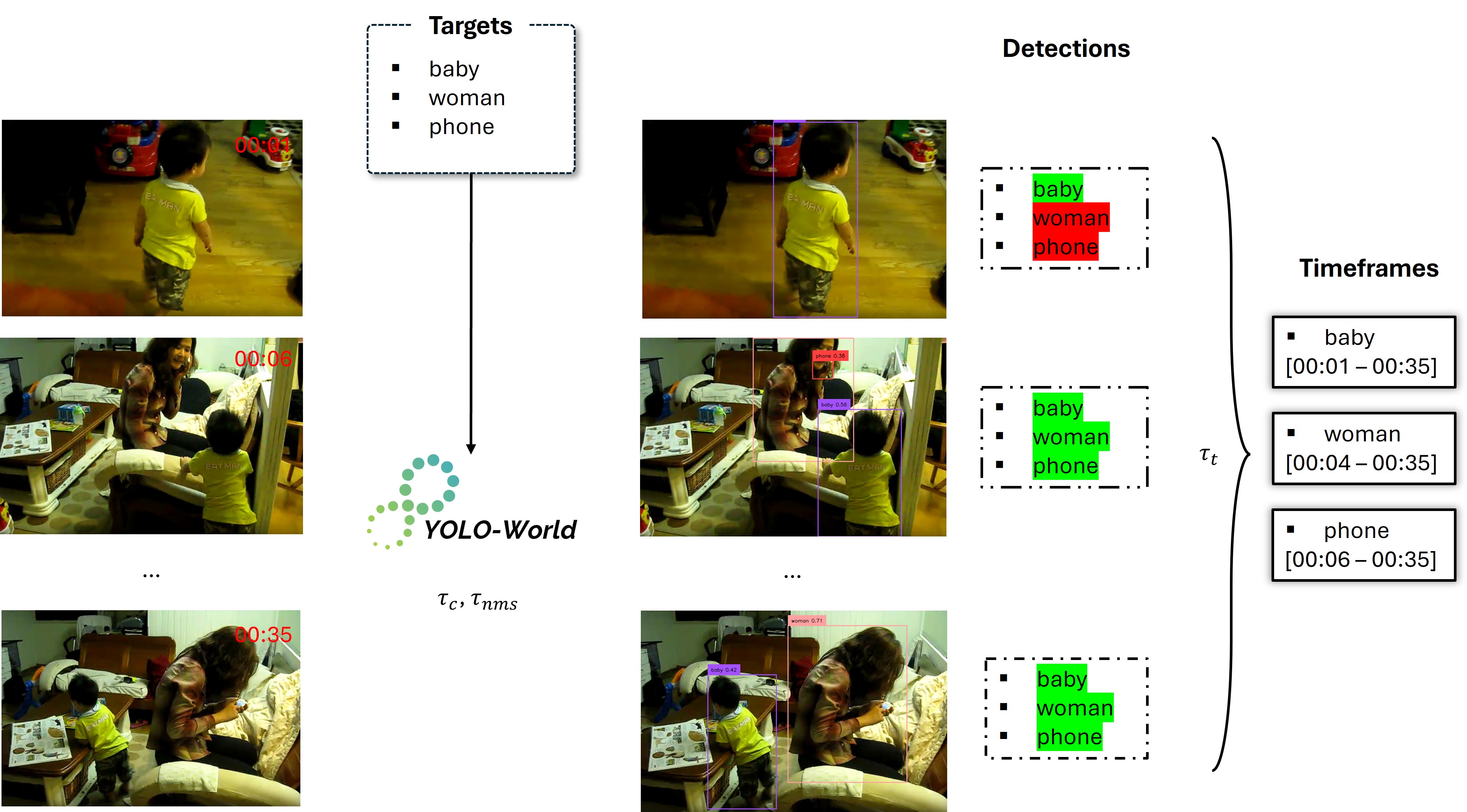} \hfill
  \caption {\textbf{A more detailed overview of the internal process of the $OG$ module.} The process begins by extracting all frames from the input video $V$. For each frame $v_i$, the YOLO-World model detects specified target classes within the frame, using predetermined confidence and NMS thresholds ($\tau_c, \tau_{nms}$). After detection, these classes are tracked across all frames to establish the exact time intervals during which they are present. If a detected object is absent from subsequent frames for a specified duration $\tau_t$, it is assumed to have exited the scene, marking the end of its appearance.}
  \label{fig:pic3}
\end{figure}

\subsection{Open-Vocabulary Object Grounding}
\label{sec:method-og}

The grounding module $OG$ (\cref{fig:pic3}) performs an object-tracking task using the open-vocabulary targets identified by the analyzer module $M_1$. This module is responsible for detecting and locating in time target objects within the video frames and is structured as follows:
\\
\textbf{1. Object Detection:} Utilizing YOLO-World \cite{yoloworld} or a similar open-vocabulary object detection model, this phase processes each video frame $v_i$ to identify instances of the relevant targets. Applying a confidence threshold to $\tau_{c}$ ensures that only credible detections are retained. Additionally, an NMS (non-maximum suppression) threshold $\tau_{nms}$ is used to filter redundant or overlapping detections, preserving the most significant object representations and avoiding overlapping noise in the results. 
\\
\textbf{2. Timeframe Extraction:} With object detections in place, this phase extracts timeframes where targets are identified, considering a time threshold $\tau_{t}$ that reduces the risk of short-lived false negatives (\eg, objects momentarily obscured or blurred). This time threshold is highly beneficial to ensure precise temporal localization so that detected objects remain consistent with the scene continuity without responding to minor, transient distortions. As a result, the grounding module outputs a set of well-defined timeframes where each target is reliably tracked. These extractions are a list of timeframes for a particular target $\{ T_1: [(t_{i0},t_{f0}), ..., (t_{in},t_{fn})], ... \}$.

These three hyperparameters ($\tau_{c}, \tau_{nms}, \tau_{t}$) were fine-tuned under a set of tests before running the benchmarks and are crucial to achieving balanced and precise object grounding having been optimized to ensure comprehensive target coverage without excessive or unreliable detections.

\subsection{CoT Judgment}
\label{sec:method-m2}

The final module $M_2$ performs in-depth reasoning and cross-validation, combining outputs from the previous modules to deliver a definitive answer to the question. This module acts first as a judge, evaluating and consolidating all prior insights through a chain-of-thought approach, and reutilizes the $M_1$ logic in case of inconsistencies, then, with the additional information, acts as a reasoner to provide a final answer:
\\
\textbf{1. Comparison judgment} First, both the $M_1$ reasoning output and the scene-segmented timeframes are merged into a single \textit{grounded context}, that contains information regarding the overall video and a first answer to be judged. This context is compared against the $OG$ \textit{grounded object} timeline. The main output from this comparison judgment is a decision of whether they are consistent and, if not, a specific reasoning of why and where they are inconsistent.
\\
\textbf{2. Question Generation:} If the outputs are effectively discrepant, with the aid of the why/where specifications of the discrepancy, and the original question $Q$ and answer options $A_{opt}$, a set of one or more clarification questions are formulated for the given timeframe, in order to obtain additional information for being able to finally validate the confidence of the answer candidate proposed by $M_1$. These clarification questions are then fed to the $M_1$'s VideoLLM instance for simple question-answering. The answers are then analyzed in a further step.
\\
\textbf{3. Answer Refinement:} Based on the clarification questions created in the previous step and their corresponding answers given by the VideoLLM, the module adjusts or refines the initial answer where necessary, ensuring that the final output reflects a coherent understanding of the video as well as consistency with both visual and semantic information.

\begin{algorithm}[t]
\caption{ViQAgent framework algorithm, after initializing the models YoloWorld, VideoLLM$_{1,2,3,4}$, LLM$_{1,2,3}$}\label{alg:viqagent}
\begin{algorithmic}[1]
\State \textbf{Hyperparameters:} $\tau_c, \tau_{nms}, \tau_t$
\State \textbf{Input:} $V, Q, [A_{opt}]$
\State $prompt \gets Q+[A_{opt}]$
\State {\color{BurntOrange} \Comment{ $M_1$ starts}}
\State $A, R_1 \gets$ VideoLLM$_{1}(V,prompt)$ \Comment{Rationale} 
\State $TC \gets$ VideoLLM$_2(V)$ \Comment{Timeframes}
\State $T \gets$ VideoLLM$_3(V, prompt)$ \Comment{ Targets }
\State $\{ T_1, T_2, ..., T_m \} \gets T$ 
\State {\color{Cyan} \Comment{$OG$ starts}}
\State {YoloWorld\texttt{.set\_classes(}$T_1, T_2, ..., T_m$\texttt{)}}
\State {$D \gets \{ T_1:\varnothing, T_2:\varnothing, ..., T_m:\varnothing \}$} \Comment{Detections}
\For{$v_i$ in $V$} 
    \State $d \gets$ YoloWorld\texttt{.detect(}$v_i, \tau_c, \tau_{nms}$\texttt{)}
    \For{$T_i$ in $T$}
        \State $D[T_i] \gets D[T_i] + d[T_i]$
    \EndFor
\EndFor
\State {$t \gets 0$}
\State {$TG \gets \{ T_1:\varnothing, T_2:\varnothing, ..., T_m:\varnothing \}$} \Comment{Timeframes}
\For{$T_i$ in $T$} 
    \State $idx \gets 0$
    \For{$d$ in $D$}
        \State $t \gets t + (V\texttt{.fps})^{-1}$
        \If{$d[T_i] \neq \varnothing$}
            \If{$idx + 1 \neq |TG[T_i]|$}
                \State $TG[T_i][idx] \gets (t,0)$
            \Else
                \State $TG[T_i][idx][1] \gets t$
            \EndIf
        \Else
            \State $\Delta t \gets t - TG[T_i][idx][1]$
            \If{$\Delta t \geq \tau_t$}
                \State $idx \gets idx + 1$
            \EndIf
        \EndIf
    \EndFor
\EndFor
\State {\color{ForestGreen} \Comment{$M_2$ starts}}
\State $\mu \gets$ LLM$_1(R_1,TC,TG)$ \Comment{Unconfidence}
\If{$\mu$}
    \State $W \gets V$\texttt{.trim}$(\mu$\texttt{.times}$)$
    \State $qs \gets\ $LLM$_2(\mu, prompt)$
    \State $as \gets \varnothing$
    \For{$q$ in $qs$}
        \State $as \gets as + $VideoLLM$_4(W,q)$
    \EndFor
    \State $A \gets\ $LLM$_3(R_1,prompt,qs,as)$
\EndIf
\end{algorithmic}
\end{algorithm}

\subsection{ViQAgent}
\label{sec:method-all}

The end-to-end ViQAgent pipeline integrates these modules in a sequential, end-to-end manner to provide an accurate, reliable answer for VideoQA tasks. The algorithm of the whole framework can be seen in \cref{alg:viqagent}.

By structuring each module’s operations this way, ViQAgent provides a robust VideoQA pipeline capable of answering complex questions in a zero-shot setting. It employs open-vocabulary grounding and modular reasoning to handle diverse visual-linguistic queries.

\section{Experiments}
\label{sec:experiments}

This section outlines the setup for benchmarking and evaluating ViQAgent's performance. We detail the datasets and metrics used, compare them against relevant baselines, and discuss modification studies that inform key hyperparameter choices in our model. Implementation details and a comprehensive analysis of our results are also provided.

\subsection{Datasets and Metrics}

We evaluated ViQAgent on four widely recognized video question answering (VideoQA) benchmarks, each representing unique video types, question formats, and challenges in the domain.
\\
\textbf{1. NExT-QA \cite{nextqa}:} This dataset tests reasoning over causal, temporal, and descriptive question types. In our experiments, we used the validation split containing 4,996 video-question pairs. Each question is close-ended, presenting five answer options, with ViQAgent tasked with selecting the correct one.
\\
\textbf{2. iVQA \cite{ivqa}:} iVQA comprises instructional video clips from the HowTo100M dataset, lasting 7-30 seconds. Each video clip includes a question and an annotated set of ground truth answers, with ViQAgent evaluated on the test set, which consists of 1,879 clips. Notably, iVQA is an open-ended VideoQA task, requiring ViQAgent to generate responses without candidate options.
\\
\textbf{3. ActivityNet-QA \cite{activitynetqa}:} This dataset includes 5,800 videos, each with ten annotated question-answer pairs, covering actions, objects, locations, and events. Similar to iVQA, ActivityNet-QA is an open-ended VideoQA dataset. For consistency with previous works, we report results on the test split, utilizing an evaluation based on a large language model (LLM) comparison between ViQAgent's response and ground-truth answers (see Appendix \ref{appx:anet-eval}).
\\
\textbf{4. EgoSchema \cite{egoschema}:} Focused on long-form, egocentric video understanding, EgoSchema contains 3-minute clips sourced from the Ego4D benchmark. It includes close-ended questions with higher complexity and length than NExT-QA's, assessing a model's ability to handle extended video content. We evaluated ViQAgent on the available 500 samples of the open-answer split, focusing solely on accuracy. This experiment, in particular, is not considered a paper contribution, as we didn't evaluate the whole benchmark, but it still provides an addition to previous results.

Across all benchmarks, accuracy was used as the primary evaluation metric, providing a direct and interpretable measure of ViQAgent’s performance on close-ended and open-ended tasks alike.

\begin{table*}
  \centering
  \begin{tabular}{ccccccc}
    \hline
    \rule[-6pt]{0pt}{18pt} Method & \rule[-6pt]{0pt}{18pt} Zero-Shot & \rule[-6pt]{0pt}{18pt} Agent & \rule[-6pt]{0pt}{18pt} Acc@C & \rule[-6pt]{0pt}{18pt} Acc@T & \rule[-6pt]{0pt}{18pt} Acc@D & \rule[-6pt]{0pt}{18pt} Acc@All \\
    \hline
    HiTeA & \xmark & \xmark & 62.4 & 58.3 & 75.6 & 63.1 \\
    LLaMa-VQA & \xmark & \xmark & 72.7 & 69.2 & 75.8 & 72.0 \\
    SeViLa (fine-tuned) & \xmark & \xmark & - & - & - & 73.8 \\
    \hline\hline
    InternVideo & \cmark & \xmark & 43.4 & 48.0 & 65.1 & 49.1 \\
    AssistGPT & \cmark & \xmark & 60.0 & 51.4 & 67.3 & 58.4 \\
    SeViLa & \cmark & \xmark & 61.3 & 61.5 & 75.6 & 63.6 \\
    \hline\hline
    ProViQ & \cmark & \cmark & - & - & - & 63.8\\
    ViperGPT+ & \cmark & \cmark & - & - & - & 64.0 \\
    JCEF & \cmark & \cmark & - & - & - & 66.7 \\
    LLoVi & \cmark & \cmark & 69.5 & 61.0 & 75.6 & 67.7 \\
    TraveLER & \cmark & \cmark & 70.0 & 60.5 & 78.2 & 68.2 \\
    MoReVQA & \cmark & \cmark & - & - & - & 69.2 \\
    VideoAgent (mem) & \cmark & \cmark & 60.0 & 76.0 & 76.5 & 70.8 \\
    VideoAgent (long) & \cmark & \cmark & 72.7 & 64.5 & 81.1 & 71.3 \\
    MotionEpic & \cmark & \cmark & 75.8 & \underline{74.6} & 83.3 & 76.0\\
    \hline
    ViQAgent (ours) & \cmark & \cmark & \textbf{82.2} & 74.5 & \textbf{86.3} & \textbf{80.4}\\
  \end{tabular}
  \caption{Results of ViQAgent against all the other state-of-the-art solutions in NExT-QA benchmark, in all the casual (C), temporal (T), and descriptive (D) subsets, as well as the overall accuracy. ViQAgent surpasses all the baselines, achieving a new state-of-the-art on VideoQA.}
  \label{tab:nextqa}
\end{table*}

\begin{table*}
  \centering
  \begin{subtable}{0.45\linewidth}
      \begin{tabular}{ccccc}
        \hline
        \rule[-6pt]{0pt}{18pt} Method & \rule[-6pt]{0pt}{18pt} Zero-Shot & \rule[-6pt]{0pt}{18pt} Agent &  \rule[-6pt]{0pt}{18pt} Acc \\
        \hline
        VideoCoCa & \xmark & \xmark & 39.0\\
        FrozenBiLM (fine-tuned) & \xmark & \xmark & 39.7\\
        \hline\hline
        FrozenBiLM & \cmark & \xmark & 27.3\\
        BLIP-2 & \cmark & \xmark & 45.8 \\
        InstructBLIP & \cmark & \xmark & 53.8 \\
        \hline\hline
        ProViQ & \cmark & \cmark & 50.7\\
        JCEF & \cmark & \cmark & 56.9 \\
        MoReVQA & \cmark & \cmark & 60.9\\
        \hline
        ViQAgent (ours) & \cmark & \cmark & \textbf{62.6}
      \end{tabular}
      \caption{iVQA results.}
      \label{tab:ivqa}
  \end{subtable}
  \hfill
  \begin{subtable}{0.45\linewidth}
      \begin{tabular}{ccccc}
        \hline
        \rule[-6pt]{0pt}{18pt} Method & \rule[-6pt]{0pt}{18pt} Zero-Shot & \rule[-6pt]{0pt}{18pt} Agent & \rule[-6pt]{0pt}{18pt} Acc \\
        \hline
        FrozenBiLM (fine-tuned) & \xmark & \xmark & 43.2 \\
        \hline\hline
        Video-ChatGPT & \cmark & \xmark & 35.2 \\
        Video-LLaVa & \cmark & \xmark & 45.3\\
        VideoChat2 & \cmark & \xmark & 49.1\\
        \hline\hline
        ViperGPT+ & \cmark & \cmark & 37.1 \\
        ProViQ & \cmark & \cmark & 42.3 \\
        JCEF & \cmark & \cmark & 43.3\\
        MoReVQA & \cmark & \cmark & 45.3\\
        MotionEpic & \cmark & \cmark & 54.6\\
        \hline
        ViQAgent (ours) & \cmark & \cmark & \textbf{59.9}
      \end{tabular}
      \caption{ActivityNet-QA results.}
      \label{tab:activitynet}
  \end{subtable}
\caption{Open-ended Question-Answering benchmark results.}
\label{tab:open-ended}
\end{table*}

\begin{table}
    \centering
      \begin{tabular}{ccccc}
        \hline
        \rule[-6pt]{0pt}{18pt} Method & \rule[-6pt]{0pt}{18pt} Agent & \rule[-6pt]{0pt}{18pt} Acc Subset & \rule[-6pt]{0pt}{18pt} Acc \\
        \hline
        SeViLa & \xmark & 25.7 & 22.7\\
        ImageViT & \xmark & 40.8 & 30.9\\
        ShortViViT & \xmark & 49.6 & 31.0\\
        InternVideo & \xmark & - & 32.1\\
        LongViViT & \xmark & 56.8 & 33.3\\
        MC-ViT-L & \xmark & 62.6 & 44.4\\
        Vamos & \xmark & - & 48.3\\
        \hline\hline
        JCEF & \cmark & - & 50.0\\
        LLoVi & \cmark & 57.6 & 50.3\\
        MoReVQA & \cmark & - & 51.7\\
        TraveLER & \cmark & - & 53.3\\
        VideoAgent (long) & \cmark & 60.2 & 54.1\\
        ProViQ & \cmark & - & 57.1\\
        
        \hline
        ViQAgent (ours) & \cmark & \textbf{67.87} & -
      \end{tabular}
    \caption{EgoSchema results.}
    \label{tab:egoschema}
\end{table}

\begin{table}
    \centering
    \begin{tabular}{cc}
    \hline
    \rule[-6pt]{0pt}{18pt} Benchmark & \rule[-6pt]{0pt}{18pt} Increase of the SotA\\
    \hline
        ActivityNet-QA & {\color{ForestGreen} +5.3\%} \\
        NExT-QA & {\color{ForestGreen} +4.4\%} \\
        iVQA & {\color{ForestGreen} +1.7\%} \\
        \hline
    \end{tabular}
    \caption{Increase on the state-of-the-art results on the benchmarks.}
    \label{tab:increase}
\end{table}

\subsection{Hyperparameter Tuning}

To refine ViQAgent's hyperparameters, we performed a series of simple modification studies on the NExT-QA dataset, specifically on each experiment's small, randomly sampled, evenly class-distributed subset of videos. We investigated the effects of varying detection thresholds in the object grounding module and the time threshold to separate appearance timeframes effectively:
\\
\textbf{Confidence Threshold.} We evaluated confidence values of 0.01, 0.05, 0.1, and 0.3 for the object detection model within YOLO-World. A default confidence of \textbf{0.05} was found optimal, balancing sensitivity to target objects with tolerance to noise in lower-resolution, motion-filled frames. This value is considerably lower than the 0.3 confidence suggested in the original YOLO-World paper \cite{yoloworld} due to the challenges posed by low-resolution and high-motion video frames in the benchmarks, which is not the case on image-level object detection. This reduction ensures that important objects are detected, accounting for YOLO-World's open-vocabulary capabilities, which can introduce variability in detection sensitivity based on the complexity of target object sets provided by the first module.
\\
\textbf{NMS Threshold.} The non-maximum suppression (NMS) threshold, controlling overlap in detections, was tested with intersection-over-union (IoU) values of 0.1 and 0.3. An IoU of \textbf{0.1} performed best, effectively reducing overlapping false positives while preserving distinct object instances in most of the cases, essential for ViQAgent’s open-vocabulary object grounding approach.
\\
\textbf{Time Threshold.} We tested temporal segmentation with 500ms, 1s, and 1.5s time thresholds. A \textbf{1.5-second} threshold was chosen because it captured consistent object presence across varying scene dynamics, particularly useful in lengthy benchmark videos, ensuring detection continuity despite video length and quality variances. However, minor changes on this threshold didn't result in significant differences.

\subsection{Results and Discussion}

ViQAgent not only surpasses prior state-of-the-art results across all evaluated benchmarks (\cref{tab:increase}) but also establishes itself as a leading choice in both zero-shot VideoQA and modular, agent-based solutions. In the close-ended NExT-QA benchmark (\cref{tab:nextqa}), ViQAgent demonstrated a significant edge, achieving a 4.4\% increase in accuracy over the previous best-performing Video-of-Though (VoT) solution: MotionEpic \cite{videoofthought_cot_reasoner}. One particularly notable finding is that our zero-shot model outperforms fine-tuned models that are optimized specifically for this benchmark, such as SeViLa \cite{SeViLa}, LLaMa-VQA \cite{LlamaVQA}, and HiTeA \cite{HiTeA}. This result highlights ViQAgent’s capability to generalize effectively without needing task-specific training, underscoring its potential as a more adaptable and resource-efficient solution for VideoQA tasks. This zero-shot superiority not only raises the benchmark for VideoQA performance but also signals the value of robust, generalized VideoLLMs in solving real-world, unseen problems without extensive fine-tuning.

Results across the open-ended benchmarks—iVQA (\cref{tab:ivqa}), ActivityNet-QA (\cref{tab:activitynet})—further validates ViQAgent's robustness and adaptability to diverse video types and question structures, showing its performance as a versatile VideoQA agent capable of handling both close-ended and open-ended tasks, while being able to surpass robust state-of-the-art agentic solutions for VideoQA: MoReVQA \cite{morevqa_modular_cot_reasoner} (and it's additional contributions JCEF and ViperGPT+), TraveLER \cite{traveler_agent_cot_planner}, ProViQ \cite{zeroshot_procedural}, MotionEpic \cite{videoofthought_cot_reasoner}, VideoAgent \cite{videoagent_memory}, and VideoAgent \cite{videoagent_long}.

Further, ViQAgent shows promising results on the partial open-answer subset of questions within the EgoSchema benchmark (\cref{tab:egoschema}), expanding its potential applicability as a flexible VideoQA model well-suited for both straightforward and complex agentic tasks.

\section{Conclusion}
\label{sec:conclusion}

In this paper, we presented ViQAgent, a novel framework that advances the field of video question answering (VideoQA) by leveraging a modular, task-specific approach tailored for zero-shot generalization across diverse VideoQA benchmarks. By employing a strategically layered architecture that integrates open-vocabulary object detection, object grounding, and chain-of-thought reasoning, ViQAgent dynamically adapts to the complexities of various video domains, enabling it to successfully tackle both close-ended and open-ended questions without requiring extensive hours of fine-tuning or training. Through extensive benchmarking on datasets such as NExT-QA, iVQA, and ActivityNet-QA, we demonstrated ViQAgent’s superior performance, significantly outperforming existing zero-shot and modular-agent approaches. Our experiments also highlight the adaptability of our framework to low-resolution, high-motion, and lengthy videos, settings that traditionally challenge VideoQA models. By pushing the boundaries of zero-shot VideoQA, setting a new standard for intelligent and efficient multi-modal comprehension in versatile video understanding.

\maketitlesupplementary

\appendix

This supplementary material provides detailed information about the ViQAgent framework implementation. It includes a case study showcasing the outputs at each stage of the framework and analyzing observations from each step (\cref{appx:example}). Additionally, the prompts and schemas used in the solution are detailed (\cref{appx:prompts}), along with the baselines and benchmark configurations (\cref{appx:baselines}). Finally, insights and implications are discussed in greater depth (\cref{appx:analysis}).

\section{Case Study}
\label{appx:example}

\begin{figure*}[t]
  \includegraphics[width=\linewidth]{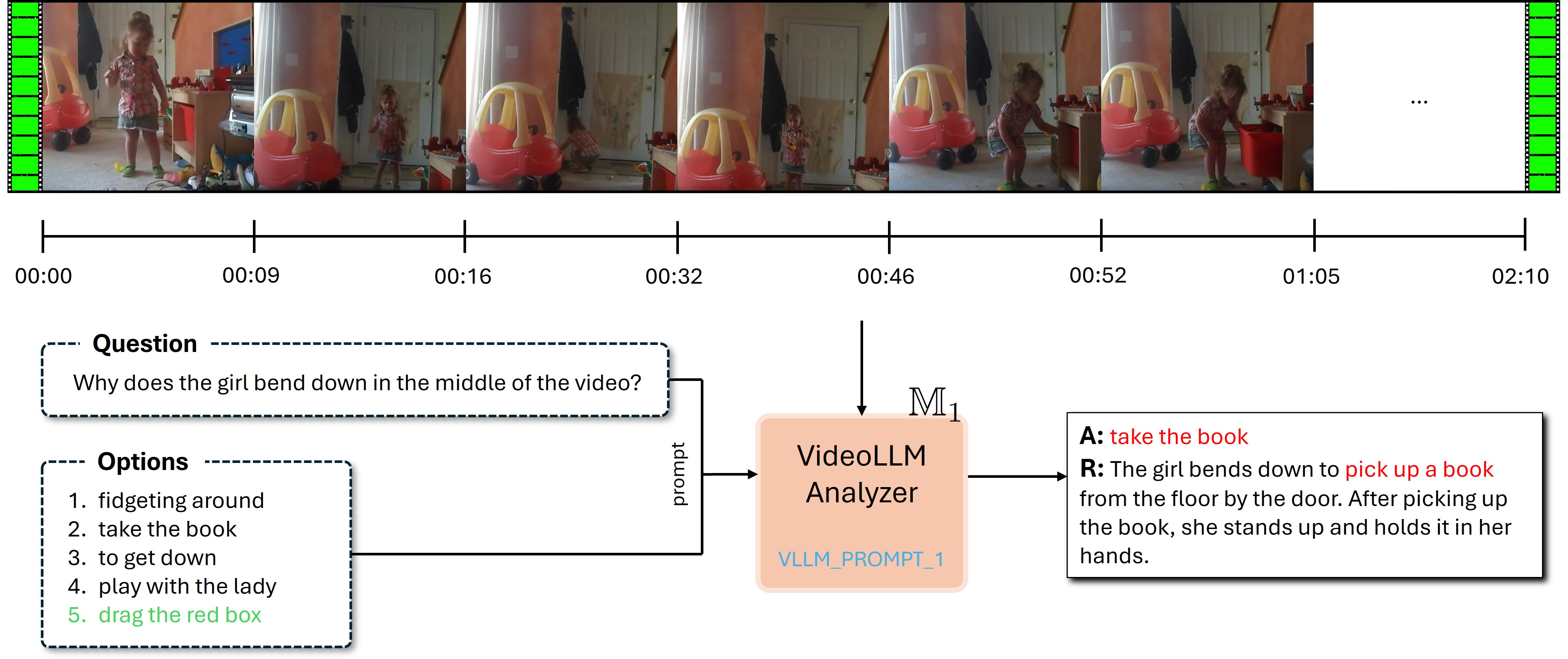} \hfill
  \caption {\textbf{VideoLLM Analyzer} {\color{BurntOrange} (VideoLLM$_1$)}: Given the full video, and the "prompt" (question plus answer options, if available), the VideoLLM Analyzer submodule provides a first-sight response with a reasoning text of why that answer is correct.}
  \label{fig:exp-m11}
\end{figure*}

\begin{figure*}[t]
  \includegraphics[width=\linewidth]{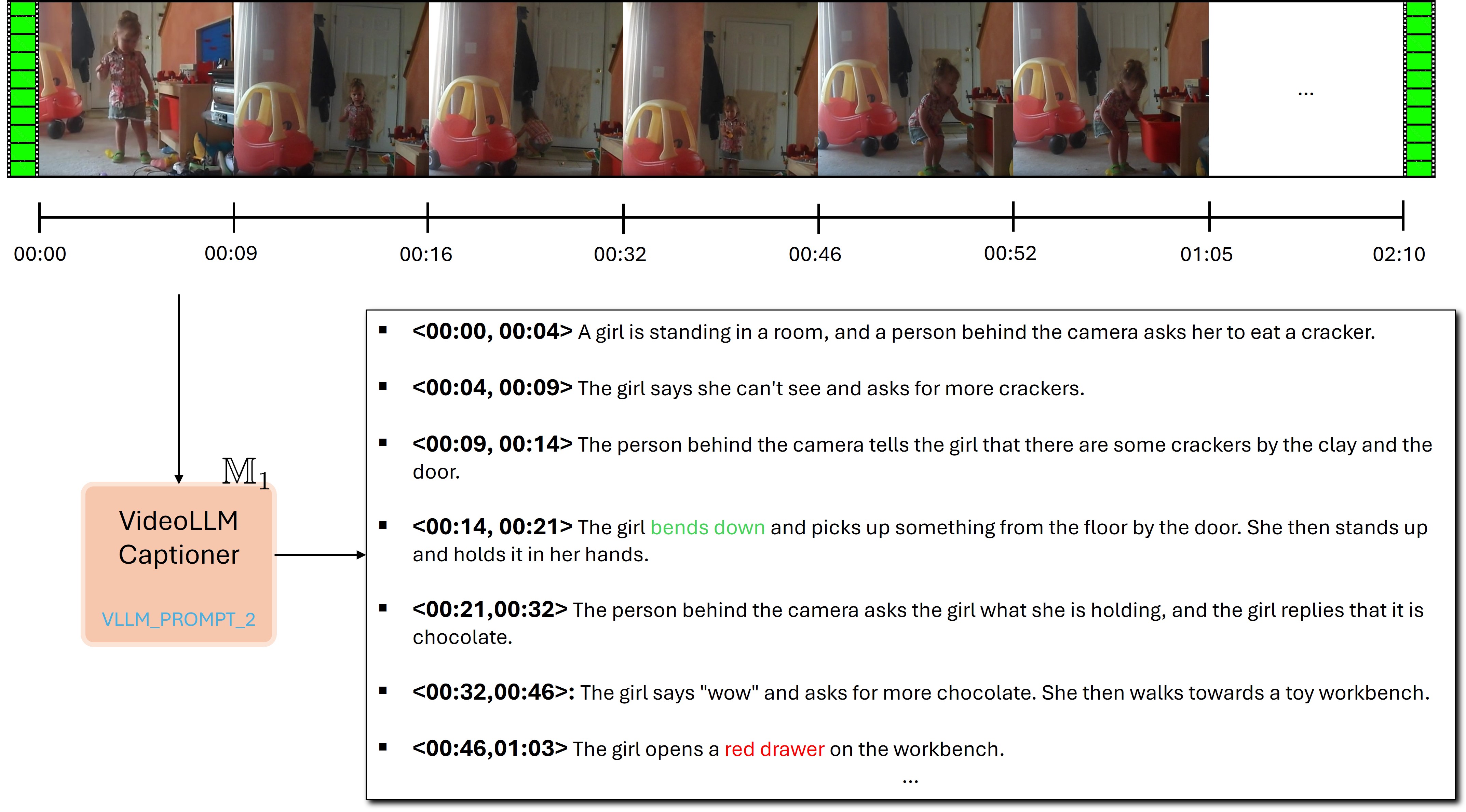} \hfill
  \caption {\textbf{VideoLLM Captioner} {\color{BurntOrange} (VideoLLM$_2$)}: Given the full video, but not the question (to avoid bias), the VideoLLM Captioner submodule provides a comprehensive set of event-separated timeframes with a description (\ie caption) of what is happening in every part of the video. This is the first grounding output, used then for comparison against YOLO-World object grounding.}
  \label{fig:exp-m12}
\end{figure*}

\begin{figure*}[t]
  \includegraphics[width=\linewidth]{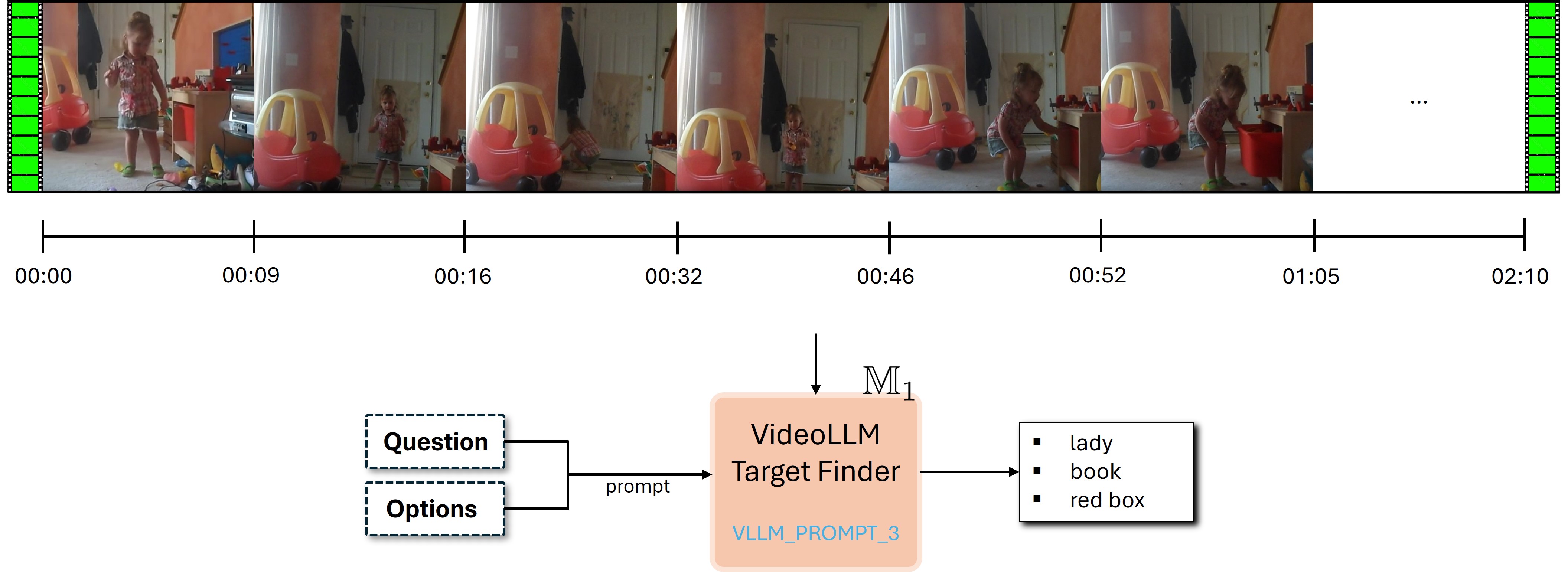} \hfill
  \caption {\textbf{VideoLLM Target Finder} {\color{BurntOrange} (VideoLLM$_3$)}: Given the full video, and the "prompt" (question plus answer options, if available), the VideoLLM Target Finder submodule is very straightforward and simple, yet effective on finding up to 4 relevant targets to identify in the video with the YOLO-World model. These objects/targets are selected based on both the video content and the relevant targets mentioned in the questions and answers.}
  \label{fig:exp-m13}
\end{figure*}

\begin{figure*}[t]
  \includegraphics[width=\linewidth]{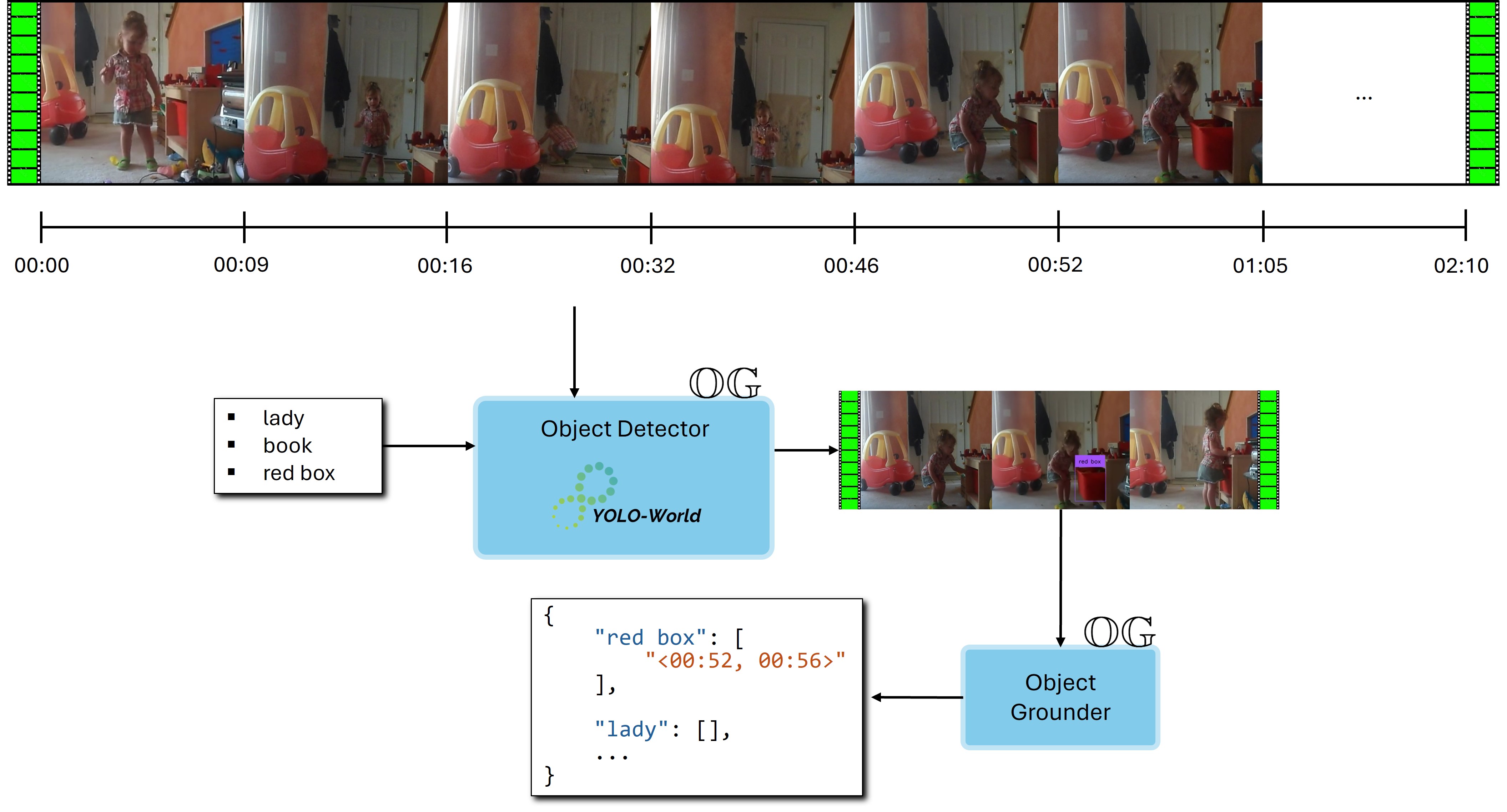} \hfill
  \caption {{\color{Cyan} ($OG$ Module)}: First, given the full video and the selected targets, the \textbf{Object Detector Model} (YOLO-World) extracts the detections from the targets in the video, which are then passed to the \textbf{Object Grounder} to consolidate into a dictionary that contains all the timeframes in which each target is detected.}
  \label{fig:exp-og}
\end{figure*}

\begin{figure*}[t]
  \includegraphics[width=\linewidth]{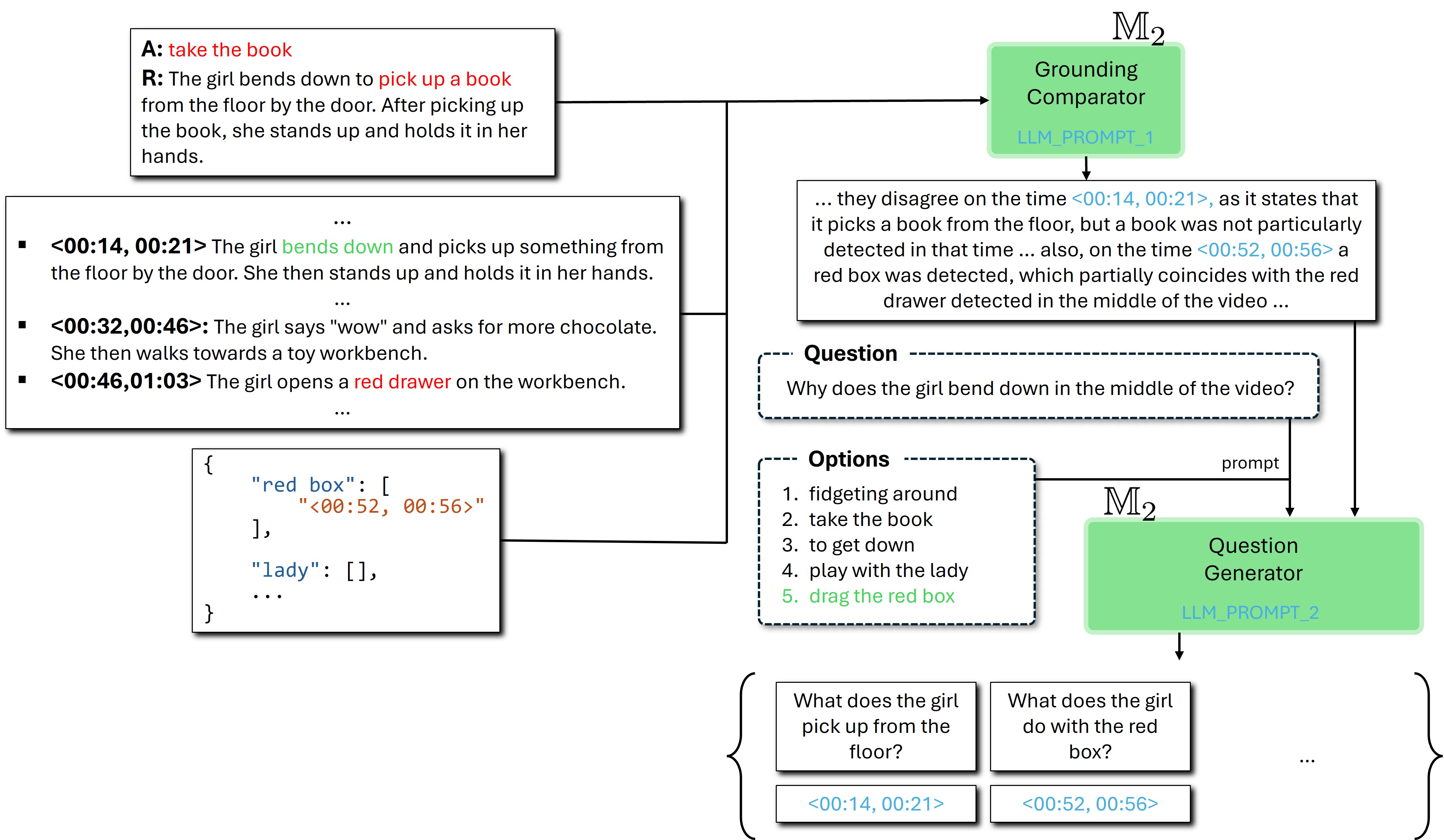} \hfill
  \caption {\textbf{LLM Grounding Comparator} {\color{ForestGreen} (LLM$_1$)}: Given the grounding output from the captioning submodule (VideoLLM$_2$), the object grounding, and the first-sight reasoning (VideoLLM$_1$), the grounding comparator determines whether there are inconsistencies or uncertain parts within them, determines, and explains in which parts are these inconsistencies, which are then fed to the \textbf{Question Generator} {\color{ForestGreen} (LLM$_2$)} along with the prompt (original question plus answer options, if available), to state up to 3 clean questions and their doubtful timeframes, to be then asked to a VideoLLM.} 
  \label{fig:exp-m21}
  
\end{figure*}

\begin{figure*}[t]
  \includegraphics[width=\linewidth]{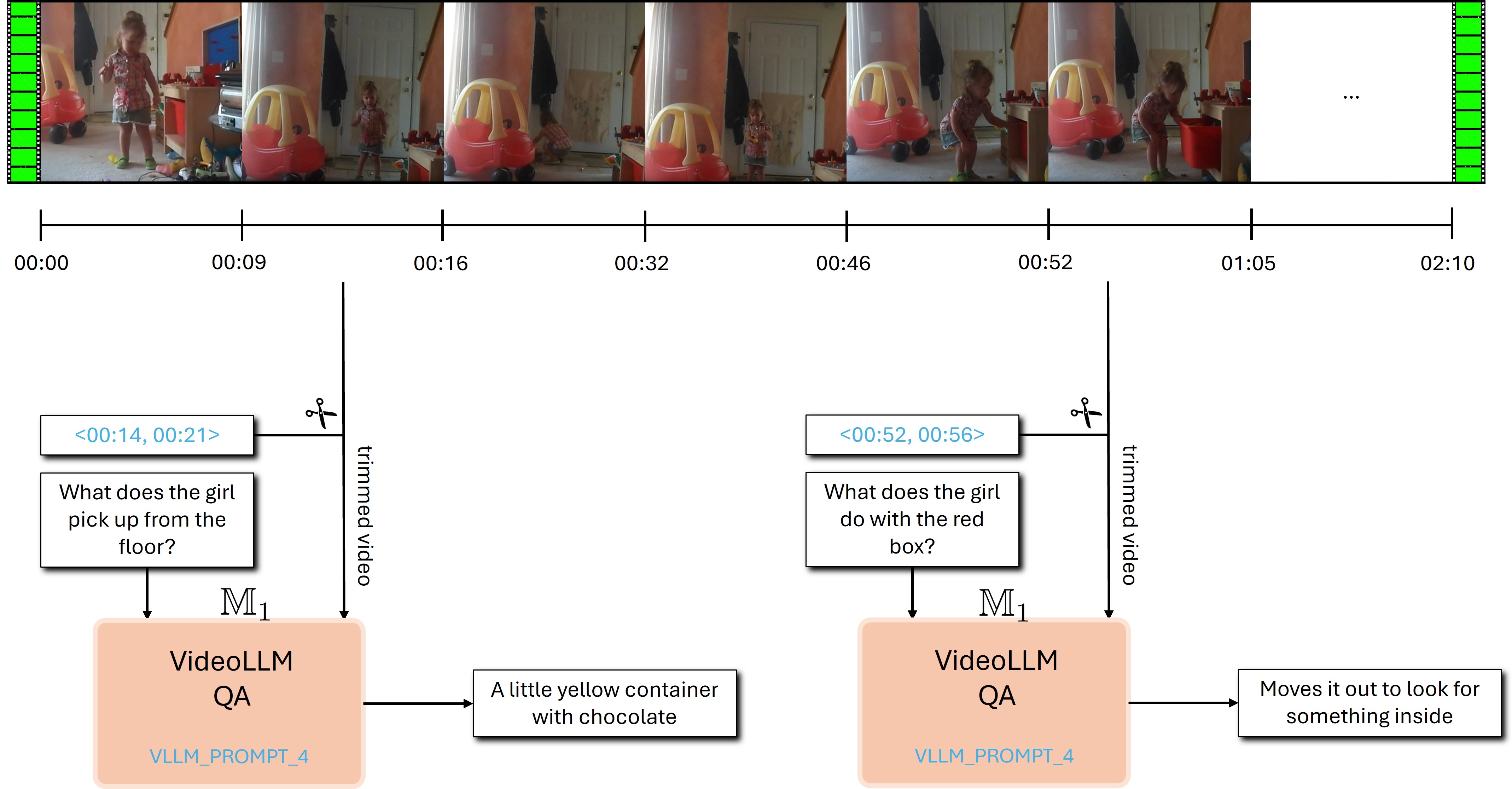} \hfill
  \caption {The stated questions from the previous step are then fed to the \textbf{VideoLLM QA} {\color{BurntOrange} (VideoLLM$_4$)} submodule to simply answer them.}
  \label{fig:exp-m22}
\end{figure*}

\begin{figure*}[t]
  \includegraphics[width=\linewidth]{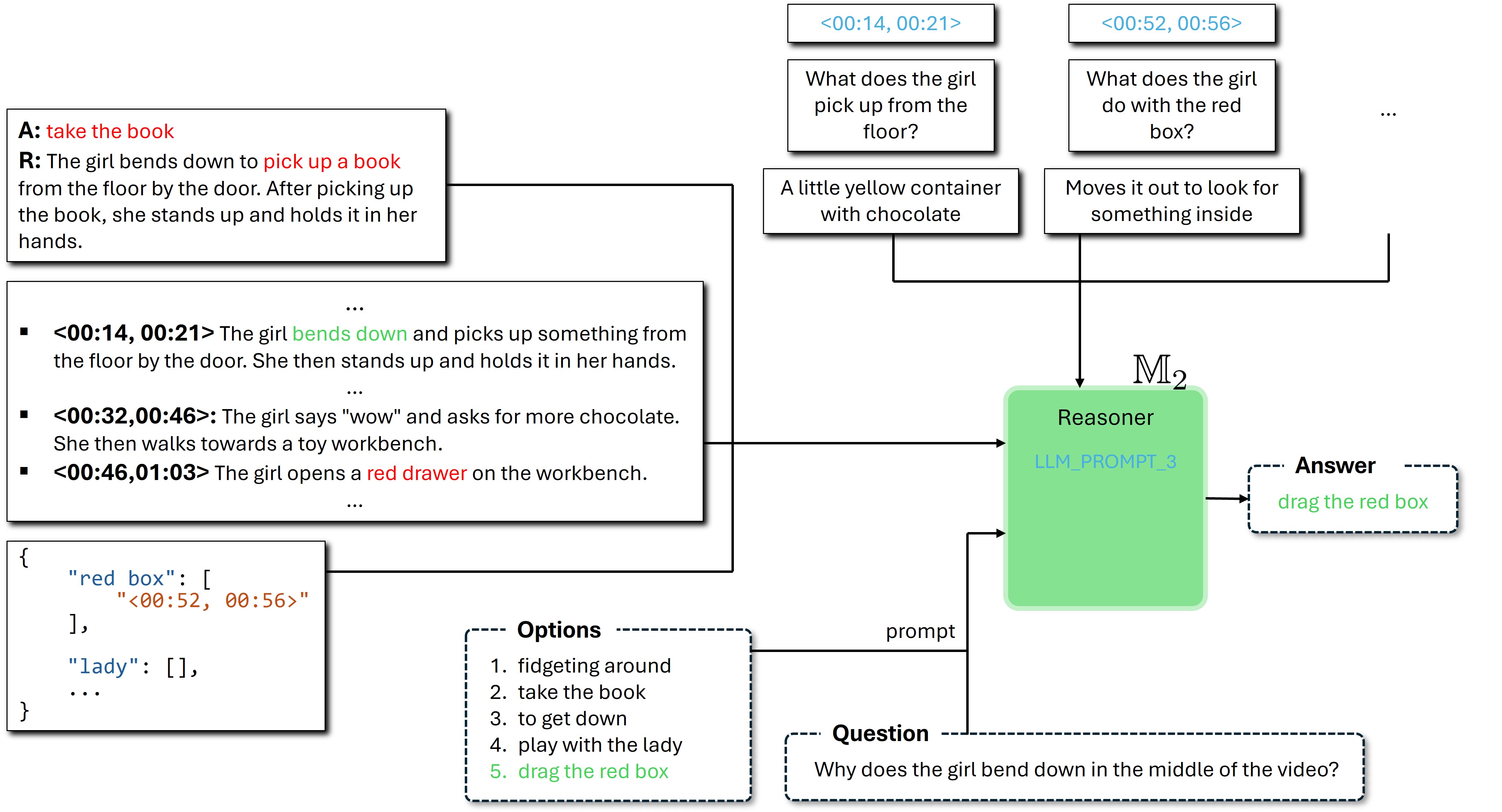} \hfill
  \caption {Once all the intermediate outputs are generated, they are fed to a \textbf{Final Reasoner} {\color{ForestGreen} (LLM$_3$)} submodule that answers the question with the new information.}
  \label{fig:exp-m23}
\end{figure*}

To illustrate the internal outputs and their overall contribution to the final answer, we present a comprehensive step-by-step case study. This example demonstrates a scenario where the VideoLLM fails to answer the question directly but provides hints that facilitate the detection of subsequent inconsistencies, ultimately leading to a correct answer. The case study is detailed in Figs. \ref{fig:exp-m11}–\ref{fig:exp-m23}. These steps align with the methodology and sequence outlined in Algorithm \ref{alg:viqagent}.

\section{Prompts and Schemas}
\label{appx:prompts}

In the ViQAgent framework, multiple LLMs are utilized in both the $M_1$ and $M_2$ modules. Module 1 is responsible for independently extracting relevant information and reasoning from the video, while Module 2 compiles and evaluates the final answer based on the information gathered.

Module 1 directly interacts with the video through VideoLLMs, whereas Module 2 leverages LLMs for reasoning using the pre-computed information without direct access to the video. The prompts and output schemas for all VideoLLM$_{1,2,3,4}$ (Module 1) and LLM$_{1,2,3}$ are presented in Tabs. \ref{tab:vllm_prompt_1} - \ref{tab:llm_schema_3}, following the procedure shown previously in the Algorithm \ref{alg:viqagent}. Notably, for VideoLLM$_1$ (\cref{tab:vllm_prompt_1}) and LLM$_3$ (\cref{tab:llm_prompt_3}), a space is left at the end. This is because these two submodules are specifically responsible for providing direct answers to the query. As the question formats depend on the specific benchmark, an additional subinstruction is appended in this section to accommodate the requirements of the respective benchmark.

\section{Baselines and Benchmarks}
\label{appx:baselines}

For each of the evaluated benchmarks, we conducted a thorough review of the most relevant state-of-the-art solutions from recent years that have reported results on these benchmarks, focusing on those that achieved notably high accuracy or are widely recognized baselines evaluated against previous work.

\textbf{NExT-QA \cite{nextqa}:} We compare against fine-tuned solutions that achieved high performance across both specific subsets (Casual, Temporal, and Descriptive) and the overall subset. The solutions include HiTeA \cite{HiTeA}, LLaMa-VQA \cite{LlamaVQA}, and a fine-tuned version of SeViLa \cite{SeViLa}. Additionally, several zero-shot solutions emerged, as NExT-QA is a significant benchmark for evaluating Video Question-Answering solutions. We classified these solutions into agentic and non-agentic categories. The non-agentic solutions include InternVideo \cite{InternVideo}, AssistGPT \cite{AssistGPT}, and SeViLa \cite{SeViLa}. The agentic solutions, which are more pertinent for comparison with our framework, include ProViQ \cite{zeroshot_procedural}, ViperGPT+, JCEF, MoReVQA (all introduced in \citet{morevqa_modular_cot_reasoner}, with ideas from \citet{visionqa_vipergpt}), LLoVi \cite{LLoVi}, TraveLER \cite{traveler_agent_cot_planner}, VideoAgent \cite{videoagent_memory}, VideoAgent \cite{videoagent_long}, and MotionEpic \cite{videoofthought_cot_reasoner} with its Video-of-Thought (VoT) framework.

\textbf{iVQA \cite{ivqa}:} For this dataset, only two fine-tuned solutions were identified, as reported in \citet{morevqa_modular_cot_reasoner}: VideoCoCa \cite{VideoCoCa} and a fine-tuned version of FrozenBiLM \cite{FrozenBiLM}. Its non-fine-tuned counterpart, alongside BLIP-2 \cite{blip2} and InstructBLIP \cite{InstructBLIP}, forms the zero-shot non-agentic baselines for comparison. Regarding the agentic solutions, which are particularly relevant to our framework, the notable baselines include ProViQ \cite{zeroshot_procedural}, JCEF, and MoReVQA \cite{morevqa_modular_cot_reasoner}.

\textbf{ActivityNet-QA \cite{activitynetqa}:} For the ActivityNet-QA benchmark, the fine-tuned group comprises only the FrozenBiLM solution \cite{FrozenBiLM}. The non-agentic zero-shot baselines include Video-ChatGPT \cite{videochatgpt}, Video-LLaVa \cite{videollava}, and VideoChat2 \cite{Videochat2}. For agentic solutions, the primary baselines are ViperGPT+, JCEF, MoReVQA \cite{morevqa_modular_cot_reasoner}, ProViQ \cite{zeroshot_procedural}, and MotionEpic \cite{videoofthought_cot_reasoner}, which employs the VoT framework.

\textbf{EgoSchema \cite{egoschema}:} Although EgoSchema is not a primary benchmark for our solution, several baselines were compared with our partial results. It is noteworthy that fine-tuning is not feasible for this dataset, as only a subset of the answers is accessible. Therefore, only non-agentic and agentic solutions were included in the evaluation. Among the non-agentic baselines, the most relevant are SeViLa \cite{SeViLa}, InternVideo \cite{InternVideo}, and Vamos \cite{Vamos}. For agentic solutions, previously established baselines were used, including LLoVi \cite{LLoVi}, TraveLER \cite{traveler_agent_cot_planner}, VideoAgent \cite{videoagent_long}, ProViQ \cite{zeroshot_procedural}, JCEF, and MoReVQA \cite{morevqa_modular_cot_reasoner}.

\subsection{ActivityNet-QA evaluation}
\label{appx:anet-eval}

ActivityNet-QA \cite{activitynetqa} open-ended benchmark provides only a single correct answer for each question, requiring LLM-based evaluation of responses, as shown in prior work \cite{gemini15, morevqa_modular_cot_reasoner, videochatgpt}. Notably, this process was not required for iVQA \cite{ivqa}, which offers a set of potential answers. For this evaluation, we used the following prompt:

{\begin{verbatim}
Evaluate whether the predicted answer
/reasoning are correct based on the real 
answer to the question. Only output 'yes'
or 'no', don't provide an explanation.

Question: {q}
Real answer: {a}
Predicted answer: {p}
Predicted reasoning: {r}
Output (yes/no): 
\end{verbatim}}

\section{Additional Analysis}
\label{appx:analysis}

As demonstrated in the benchmark results, the ViQAgent framework shows significant promise for the Video Question Answering task and, in general, tool-using agents, establishing a new state-of-the-art by integrating VideoLLMs, LLMs, and computer vision foundation models to address questions based on video content. However, it is important to acknowledge that, as LLMs currently cannot be seeded, certain random factors in their output remain uncontrollable, even though all experiments were conducted with a temperature parameter set to $0.0$. Furthermore, the ViQAgent framework proves most effective when the object or target of the query is visibly present in the video. In cases where the YOLO-World model cannot detect the object, confusion may arise. Nevertheless, the framework is robust enough to mitigate this confusion through the CoT judge module ($M_2$), resulting in improved performance while preserving the primary advantages of Zero-Shot VideoLLM-based question answering.

\ 
\clearpage



\ 
\clearpage


\begin{table*}[t]
\caption{VideoLLM Analyzer Prompt (VideoLLM$_1$)}
\label{tab:vllm_prompt_1}
\begin{promptbox}[label={VLLM_PROMPT_1}]{VideoLLM Analyzer}
\begin{verbatim}
VLLM_PROMPT_1 = """
Based on the provided video, select or provide the correct answer for the user 
question. Break down your reasoning into clear, logical steps, and arrive at 
the most accurate answer.

To ensure accuracy, follow this step-by-step reasoning process:
1. Restate or reframe the question for clarity.
2. Consider key events, actions, or objects relevant to the question.
3. If answer options are provided, assess each option in relation to the 
video's content. If no options are given, logically derive an answer.
4. Provide a clear and concise response based on your reasoning.

You must provide the index of the selected answer or the answer itself, and a 
brief explanation of your reasoning.

"""
\end{verbatim}
\end{promptbox}
\end{table*}

\begin{table*}[t]
\caption{VideoLLM Analyzer Schema (VideoLLM$_1$)}
\label{tab:vllm_schema_1}
\begin{mybox}[label={VLLM_SCHEMA_1}]{Output Schema for VideoLLM Analyzer}
\begin{verbatim}
VLLM_SCHEMA_1 = {
    "type": "object",
    "properties": {
        "reasoning": { "type": "string" },
        "answer": { "type": "string" }
    }
}
\end{verbatim}
\end{mybox}
\end{table*}

\begin{table*}[t]
\caption{VideoLLM Captioner Prompt (VideoLLM$_2$)}
\label{tab:vllm_prompt_2}
\begin{promptbox}[label={VLLM_PROMPT_2}]{VideoLLM Captioner}
\begin{verbatim}
VLLM_PROMPT_2 = """
Based on the provided video and the given question (and answer options if 
available), capture a list of the main timeframes in the video in the format 
<<mm0:ss0,mm1:ss1>>: {description}, where 'description' is a detailed 
description of what is happening in that particular timeframe.

Follow these steps to generate your response:
1. Carefully analyze the question and the video content to identify the key 
events or actions that are relevant to the question.
2. Identify key events, actions, or transitions that represent meaningful 
changes or notable moments in the video.
3. Break the video into distinct timeframes where these events occur.
4. For each identified timeframe, provide a clear, detailed description of the 
action or scene in that segment.
5. Ensure that each description is specific, concise, and accurately reflects 
the action within the timeframe.
"""
\end{verbatim}
\end{promptbox}
\end{table*}

\begin{table*}[t]
\caption{VideoLLM Captioner Schema (VideoLLM$_2$)}
\label{tab:vllm_schema_2}
\begin{mybox}[label={VLLM_SCHEMA_2}]{Output Schema for VideoLLM Captioner}
\begin{verbatim}
VLLM_SCHEMA_2 = {
    "type": "object",
    "properties": {
        "timeframes": {
            "type": "array",
            "items": { "type": "string" }
        }
    }
}
\end{verbatim}
\end{mybox}
\end{table*}

\begin{table*}[t]
\caption{VideoLLM Target Finder Prompt (VideoLLM$_3$)}
\label{tab:vllm_prompt_3}
\begin{promptbox}[label={VLLM_PROMPT_3}]{VideoLLM Target Finder}
\begin{verbatim}
VLLM_PROMPT_3 = """
Based on the provided video and the given question (and answer options if 
available), your task is to capture a **list of objects/targets** that are 
involved in the video and are relevant to the question. These targets will 
be used for object detection and grounding via a YOLO model. Please follow 
these steps:

1. Understand the question and its context within the video, along with any 
answer options provided.
2. Focus on the most relevant objects or targets that are involved in the 
video's key actions or scenes. Ensure that these targets directly relate to 
the question.
3. Choose no more than 4 targets, ideally 3 or fewer. Consider only the 
objects that are clearly present and essential to answering the question, 
and that are not too complex to identify (not too large as well), but not 
too general for the particular video.
4. Ensure that the targets are also directly related to the answer options, 
if provided.
5. Provide a short list of targets, ensuring each description is clear and 
relevant (e.g., 'player in white outfit', 'spoon', etc.).
"""
\end{verbatim}
\end{promptbox}
\end{table*}

\begin{table*}[t]
\caption{VideoLLM Target Finder Schema (VideoLLM$_3$)}
\label{tab:vllm_schema_3}
\begin{mybox}[label={VLLM_SCHEMA_3}]{Output Schema for VideoLLM Target Finder}
\begin{verbatim}
VLLM_SCHEMA_3 = {
    "type": "object",
    "properties": {
        "targets": {
            "type": "array",
            "items": { "type": "string" }
        }
    }
}
\end{verbatim}
\end{mybox}
\end{table*}

\begin{table*}[t]
\caption{VideoLLM QA Prompt (VideoLLM$_4$)}
\label{tab:vllm_prompt_4}
\begin{promptbox}[label={VLLM_PROMPT_4}]{VideoLLM QA}
\begin{verbatim}
VLLM_PROMPT_4 = """
Based on the provided video, answer the user question in the VERY SPECIFIC 
given timeframe.

Only provide the final, concise answer, directly related to the question. 
Base your answer ONLY on the information in the video, and do not add any 
information. If the answer is not present in the video, state 'unanswerable'. 
For example, if the question is 'What color is the car?', and the car is not 
shown in the video timeframe, the answer should be 'unanswerable'.
"""
\end{verbatim}
\end{promptbox}
\end{table*}

\begin{table*}[t]
\caption{VideoLLM QA Schema (VideoLLM$_4$)}
\label{tab:vllm_schema_4}
\begin{mybox}[label={VLLM_SCHEMA_4}]{Output Schema for VideoLLM QA}
\begin{verbatim}
VLLM_SCHEMA_4 = {
    "type": "object",
    "properties": {
        "answer": { "type": "string" }
    }
}
\end{verbatim}
\end{mybox}
\end{table*}

\begin{table*}[t]
\caption{LLM Grounding Comparator Prompt (LLM$_1$)}
\label{tab:llm_prompt_1}
\begin{promptbox}[label={LLM_PROMPT_1}]{LLM Grounding Comparator}
\begin{verbatim}
LLM_PROMPT_1 = """
You will be provided with reasoning for an answer to a question, along with 
two grounding pieces of information:
1. **VideoLLM-extracted grounding captions**: These describe the key events 
and timeframes within the video (e.g., <<mm0:ss0,mm1:ss1>>: {description}).
2. **YOLO object grounding**: This identifies the specific objects/targets 
and their appearances in different video timeframes.

Your task is to analyze if there is any disagreement between the grounding 
information (both the captions and object grounding) and the reasoning for 
the answer. Disagreements may occur if the reasoning implies events or objects 
appearing in timeframes that are inconsistent with the grounding.

Please output a "disagree" boolean indicating if there is any disagreement at 
all, and a detailed but concise explanation of the specific timeframes where 
the grounding information does not align with the reasoning. Only include 
timeframes where discrepancies occur, and keep the explanation short but clear. 
If no disagreement is found, simply explain that there is no disagreement.

Disagreements should be highlighted by timeframe (<<mm0:ss0,mm1:ss1>>) and why 
the reasoning conflicts with the provided grounding information.
"""
\end{verbatim}
\end{promptbox}
\end{table*}

\begin{table*}[t]
\caption{LLM Grounding Comparator Schema (LLM$_1$)}
\label{tab:llm_schema_1}
\begin{mybox}[label={LLM_SCHEMA_1}]{Output Schema for LLM Grounding Comparator}
\begin{verbatim}
LLM_SCHEMA_1 = {
    "type": "object",
    "properties": {
        "reasoning": { "type": "string" },
        "disagree": { "type": "boolean" }
    }
}
\end{verbatim}
\end{mybox}
\end{table*}

\begin{table*}[t]
\caption{LLM Question Generator Prompt (LLM$_2$)}
\label{tab:llm_prompt_2}
\begin{promptbox}[label={LLM_PROMPT_2}]{LLM Question Generator}
\begin{verbatim}
LLM_PROMPT_2 = """
You will be provided the following:
1. A question (and answer options if available) related to a video.
2. A text explaining the set of discrepancies found in previous studies of the 
video. These indicate specific timeframes in the video where the grounding 
information does not align with the reasoning. These timeframes and the reasons 
for the discrepancies are provided.

Your task is to generate a set of up to 3 concise questions to ask a VideoLLM 
to clarify and provide a more grounded, precise answer. The goal is to resolve 
the discrepancies and improve the grounding for the question at hand.

- Each question should focus on a specific timeframe where a discrepancy was 
found.
- Each question should be concise and relevant to the timeframe, and 
particularly relevant to answer the question.
- Ensure that each question includes the timeframe where the clarification 
is needed, formatted as <<mm0:ss0,mm1:ss1>>.
- The timeframe must be very precise in time, covering only the specific 
segment where the discrepancy occurred.
- Do not include any unnecessary details, just the specific query for 
clarification.
- If there are not CONSIDERABLE discrepancies, you may return an empty list!

Generate between 0 and up to 3 questions based on the discrepancies identified.
"""
\end{verbatim}
\end{promptbox}
\end{table*}

\begin{table*}[t]
\caption{LLM Question Generator Schema (LLM$_2$)}
\label{tab:llm_schema_2}
\begin{mybox}[label={LLM_SCHEMA_2}]{Output Schema for LLM Question Generator}
\begin{verbatim}
LLM_SCHEMA_2 = {
    "type": "object",
    "properties": {
        "questions": {
            "type": "array",
            "items": { "type": "string" }
        }
    }
}
\end{verbatim}
\end{mybox}
\end{table*}

\begin{table*}[t]
\caption{LLM Final Reasoner Prompt (LLM$_3$)}
\label{tab:llm_prompt_3}
\begin{promptbox}[label={LLM_PROMPT_3}]{LLM Final Reasoner}
\begin{verbatim}
LLM_PROMPT_3 = """
You will be provided the following:
1. A question (and answer options if available) related to a video.
2. An initial reasoning made for a possible answer, along with an 
explanation of why it was chosen. This reasoning was done BEFORE 
knowing the grounding information, and clarification questions.
3. The **grounding information**:
    - **VideoLLM grounding**: Timeframes and event descriptions from the 
    video.
    - **YOLO object grounding**: Objects/targets identified in the video 
    and their corresponding appearing timeframes.
4. A set of clarification questions asked about discrepancies in the 
grounding, and their responses.

Your task is to:
1. Analyze all the provided information and reasoning.
2. Select or provide the correct answer for the user question, based on the 
new clarifications from the questions and grounding data. 
3. Provide the final, most accurate specific answer, as well as a reasoning 
for it.

Remember to stick to the information provided, and ensure that your answer 
is accurate and well-supported by the grounding information and reasoning 
provided. If none of the answer options are correct, select the most 
appropiate based on the new information and reasoning.

"""
\end{verbatim}
\end{promptbox}
\end{table*}

\begin{table*}[t]
\caption{LLM Final Reasoner Schema (LLM$_3$)}
\label{tab:llm_schema_3}
\begin{mybox}[label={LLM_SCHEMA_3}]{Output Schema for LLM Final Reasoner}
\begin{verbatim}
LLM_SCHEMA_3 = { 
    "type": "object",
    "properties": {
        "reasoning": { "type": "string" },
        "answer": { "type": "string" }
    }
}
\end{verbatim}
\end{mybox}
\end{table*}


\end{document}